\documentclass[conference]{IEEEtran}
\IEEEoverridecommandlockouts
\usepackage{amsmath,amssymb,amsfonts}
\usepackage{cases}
\usepackage{algorithmic}
\usepackage{graphicx}
\usepackage{textcomp}
\usepackage{xcolor}
\usepackage{multirow}
\usepackage{diagbox}
\usepackage{hyperref}

\def\BibTeX{{\rm B\kern-.05em{\sc i\kern-.025em b}\kern-.08em
    T\kern-.1667em\lower.7ex\hbox{E}\kern-.125emX}}

\begin{document}


\title{Pose2Drone: A Skeleton-Pose-based Framework for Human-Drone Interaction}

\author{
\IEEEauthorblockN{Zdravko Marinov\IEEEauthorrefmark{1}, Stanka Vasileva\IEEEauthorrefmark{1}, Qing Wang\IEEEauthorrefmark{1}, Constantin Seibold\IEEEauthorrefmark{2}, Jiaming Zhang\IEEEauthorrefmark{2} and Rainer Stiefelhagen\IEEEauthorrefmark{2}}
\IEEEauthorblockA{Institute for Anthropomatics and Robotics, Karlsruhe Institute of Technology, Germany}
\IEEEauthorblockA{\IEEEauthorrefmark{1}\{firstname.lastname\}@student.kit.edu, \IEEEauthorrefmark{2}\{firstname.lastname\}@kit.edu}

\thanks{This work was supported in part through the AccessibleMaps project by the Federal Ministry of Labor and Social Affairs (BMAS) under the Grant No. 01KM151112, in part by the University of Excellence through the ``KIT Future Fields'' project.}
\thanks{Code is available at: \href{https://github.com/Zrrr1997/Pose2Drone}{https://github.com/Zrrr1997/Pose2Drone}}
}

\maketitle

\begin{abstract}
 Drones have become a common tool, which is utilized in many tasks such as aerial photography, surveillance, and delivery. However, operating a drone requires more and more interaction with the user. A natural and safe method for Human-Drone Interaction (HDI) is using gestures. In this paper, we introduce an HDI framework building upon skeleton-based pose estimation. Our framework provides the functionality to control the movement of the drone with simple arm gestures and to follow the user while keeping a safe distance. We also propose a monocular distance estimation method, which is entirely based on image features and does not require any additional depth sensors. To perform comprehensive experiments and quantitative analysis, we create a customized testing dataset. The experiments indicate that our HDI framework can achieve an average of 93.5\% accuracy in the recognition of 11 common gestures. The code will be made publicly available to foster future research.
\end{abstract}


\section{Introduction}


As drones, or Unmanned  Ariel  Vehicles  (UAVs), have become increasingly more accessible, the interest in integrating them into different aspects of our lives is continuously growing. As such, drones are not only used for photography and filming~\cite{drone-cinematography-1, drone-cinematography-2}, or shipping and delivery\cite{drone-delivery}, but they are also utilized for navigating visually impaired persons\cite{drone-navigator}. For this task, assisting drones are often operated via a remote control\cite{remote-control}. This hinders the dynamic interaction as one requires either excessive human input\cite{HDI-survey}, outlaying a path via markers\cite{follow-route} or an integrated GPS-module\cite{drone-and-me}. This issue becomes especially apparent when one wants to translate interactive drones into unseen environments. 
To tackle this problem, one needs a method that is robust to domain changes such as switching between indoor and outdoor environments and requires a minimally complex user input. 
Thus, we propose a framework, which is based on a skeleton pose prediction of the human pose. This enables the drone to recognize gestures and react to them in real-time while being robust to domain changes.
The drone's movement can be controlled by a predefined set of arm gestures, making it possible to maneuver the UAV intuitively by any user without the need for any prior technical knowledge. We also provide a face-following mode, in which the drone follows the user's face and maintains a safe distance.

\begin{figure}[t]
    \begin{center}
       \includegraphics[width=1.0\linewidth, keepaspectratio]{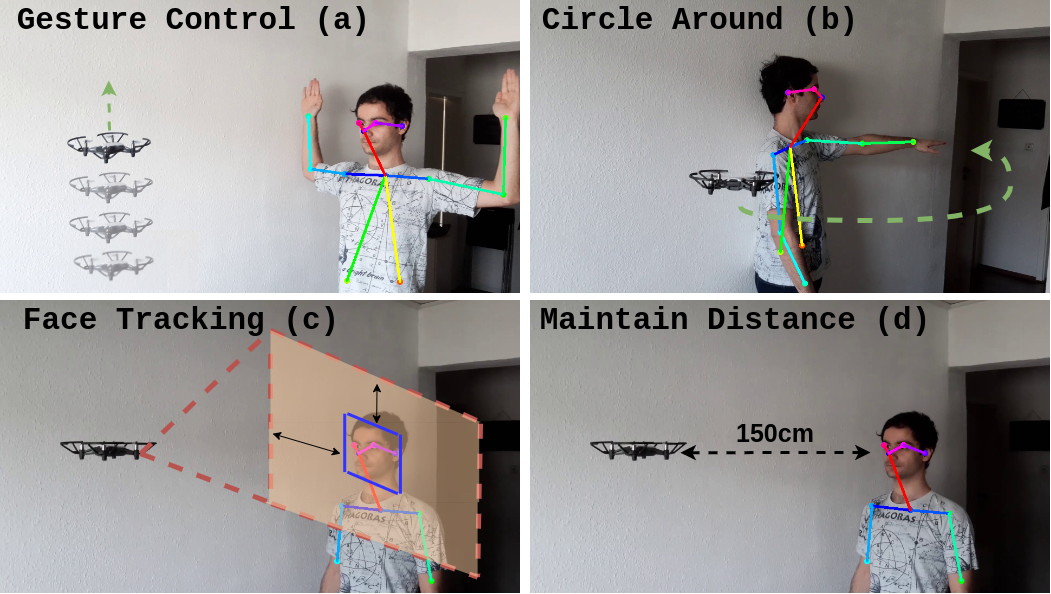}
    \end{center}
       \caption{Examples of our framework's functionality: (a) A frontal gesture command for an upward movement, (b) A gesture which commands the drone to come in front of the user, (c) Face tracking keeps the user's face in the center of the frame, and (d) The drone maintains a distance to the user.}
    \label{fig:drone_in_use}
\vskip -3ex
\end{figure}


\textit{Contributions.} In this paper, we propose the first, to our knowledge, existing skeleton-pose-based drone control framework. After a specific user registers as main applicant for the drone via facial recognition features, the user's control is characterized by two different modes for the drone's movement. The first mode, Face Following (FF), allows the drone to track the user's face while keeping a certain distance from him as seen in Fig. \ref{fig:drone_in_use}(c) and Fig. \ref{fig:drone_in_use}(d). The second mode, Gesture Control (GC), allows commanding the drone with arm gestures, in which the drone performs a specific action according to a predefined gesture-action pair. The functionality of our framework can be seen in Fig. \ref{fig:drone_in_use}.

\begin{figure*}[t]
\begin{center}
    \includegraphics[width= 1.0\linewidth, keepaspectratio]{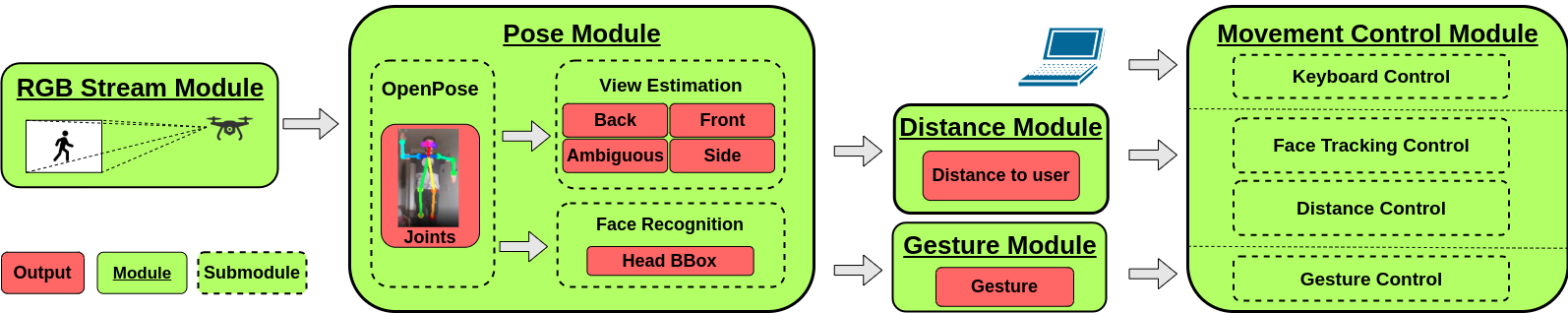}
\end{center}
   \caption{Our framework's pipeline. Each green box represents a ROS Node. The arrows show the ROS Topics and their corresponding subscribers and publishers. The outputs, indicated in red, are the data which is published in the ROS Topics. All modules are run on a Linux machine.}
\label{fig:pipeline}
\vskip -3ex
\end{figure*}

\section{Related Work}
A common method to interact with drones is \textit{e.g.} via a remote control\cite{remote-control}. This way, drone operators can maneuver drones with precision and create highly synchronized drone swarms\cite{HDI-survey}. However, controlling drones in this manner demands professional prior training, which complicates the interaction for regular users and this type of HDI often requires setting a fixed route or executing a series of predefined movements, which limits the natural interaction with the drone\cite{HDI-survey}. In contrast, while our HDI framework can integrate remote control via keyboard inputs, it enables any face-registered user to safely and intuitively manipulate the drone, utilizing the FF and GC modes.

Gestures are the most commonly used alternative modality for HDI\cite{HDI-survey} and are well-perceived\cite{drone-and-me}. Some previous approaches use depth cameras in addition to an RGB stream to determine the visibility, position, and shape of the arms and hands and recognize a gesture\cite{droneio, kinect-based}. Other methods have been proposed based only on an onboard RGB camera\cite{monocular-gestures, monocular-gestures-2}. Such methods extract visual features from the image to recognize the gesture without requiring additional depth sensors. Cauchard et al.\cite{droneio} use hand gestures for navigating a projected user interface menu and selecting actions for the drone to perform. Another usage of hand gestures is to issue movement commands directly with the gesture, \textit{e.g.}, by pointing at a direction\cite{monocular-gestures-2}. However, traditional image-based gesture recognition approaches suffer from sensitivity to occlusion and illumination variation\cite{visual-gestures-limitations}. Thus, we propose a skeleton-pose-based framework for gesture recognition, which is shown to be significantly robust to both issues.


\section{Proposed Framework}
Our framework is displayed in Fig. \ref{fig:pipeline}. Its pipeline consists of five modules, which are implemented as Robotic Operating System (ROS)\cite{ros} Nodes and communicate with each other via ROS Topics. ROS Nodes can subscribe and publish to a topic, i.e., they can receive input data and send output data via the topic. This is indicated by the gray arrows in Fig. \ref{fig:pipeline}.
\subsection{RGB Stream Module}

The RGB stream module serves to provide images, captured by the drone's camera, to the pose module. The frames are retrieved via a UDP connection with the drone and are forwarded by maintaining a predefined frame rate.  In addition, this is the only point, where frames are being transferred via ROS, which reduces the transmission overhead in the rest of the framework.

\subsection{Pose Module}
The pose module receives images from the RGB stream modules, which are used to estimate the joint locations $\text{\textbf{j}}_i = (j_{i}^x, j_{i}^y) \in\mathbb{R}^2,i\in\{0,.., 17\} $ of the user in the image with OpenPose\cite{open-pose}.
 For our framework, we only consider a subset of all joints. These are labeled with their indices $i$ in Fig.~\ref{fig:all_gestures} in the "Cheese" gesture. These locations are used as features for the view estimation and face recognition submodules as indicated by the gray arrows inside the pose module in Fig.~\ref{fig:pipeline}. We further define the distance $d(i,k)$ between two joint locations $\text{\textbf{j}}_i$ and $\text{\textbf{j}}_k$ as their 2D Euclidean distance:
 \begin{equation}
     d(i,k) = ||\text{\textbf{j}}_i - \text{\textbf{j}}_k||
 \end{equation}

\subsubsection{View Estimation}
The view estimation submodule uses the joint locations $\text{\textbf{j}}_i$ to classify the orientation of the person into the view classes $v_i\in\{v_{\text{Front}}, v_{\text{Side}}, v_{\text{Back}}, v_{\text{Ambiguous}}\}$. It is achieved by a purely geometrical approach by estimating the angle of the shoulder axis with the image plane. Since OpenPose's joint locations are 2D, the real 3D angle with the image plane can only be approximated. 
We accomplish this by comparing the $\gamma$-weighted nose-to-neck distance $d(0, 1)$ with the shoulder width $d(2,5)$, where $\gamma$ is a hyperparameter. We also differentiate between the frontal and back view with the help of the relative positions of the ears to each other.
    \begin{subnumcases}{v_i =}
         v_{\text{Side}}, & if $d(2,5) \leq \gamma \cdot d(0,1)$ \label{side_condition}\\
         v_{\text{Front}}, & if $d(2,5) > \gamma \cdot d(0, 1) \land j_{16}^x \leq j_{17}^x$ \label{front_condition} \\
         v_{\text{Back}}, & if $d(2,5) > \gamma \cdot d(0, 1) \land j_{16}^x > j_{17}^x$ 
        \label{back_condition} \\
         v_{\text{Ambiguous}}, & else
         \label{ambiguous_condition}
    \end{subnumcases}

The motive behind this method is that the nose-to-neck distance is significantly independent of the person's orientation, whereas the shoulder width becomes smaller when the person is standing sideways. We also only compute the ratio between the two distances, which makes this approach viable at any distance from the camera.

If \eqref{side_condition} does not hold, we check whether the left ear is on the left side with respect to right ear and classify the view as $v_{\text{Front}}$. Similarly, if it is on the right side, the view is classified as $v_{\text{Back}}$. However, if all \eqref{side_condition}, \eqref{front_condition}, and \eqref{back_condition} do not hold, the view class is $v_{\text{Ambiguous}}$. We introduced this class so that the definitions of the other view classes are more strict and more consistent. The modularity of our framework also allows to use a deep learning method, but our evaluations show that our geometric method is sufficient.

\subsubsection{Face Recognition}

The face recognition ensures that only one concrete person's gestures influence the drone and only his face is followed. Other people's gestures and faces in the frame are ignored. There is no conventional face detection in this module. Instead, we use the bounding box of the convex hull of the head joint locations  (\textit{i.e.}, ears, eyes, nose, and neck). To capture the whole head, the bounding box is extended by a fixed margin across the width and height. This method makes it possible to infer the user's head bounding box from all views, even where there is no apparent face on the frame, \textit{e.g.}, from the back view.

Afterward, the image crop from the bounding box is forwarded to the face recognition submodule, where we use a standard face recognition framework (see Subsection \ref{subsec:implementation}). This method requires a predefined template database images of faces and searches for the best match to the input bounding box. If the user's face is not matched in the current frame, the closest head bounding box to the last matched face is taken and tracked.
%

\subsection{Gesture Module}
The goal of the gesture module is to recognize arm gestures from a user, which could be either frontal or side gestures. The module receives the joint locations $\text{\textbf{j}}_i$ of the user from the OpenPose module as well as the estimated view class $v_i$. The gesture module determines whether a gesture is present and if the same gesture is recognized within $N$ frames, then a corresponding gesture control command is sent to the movement control, where $N$ is a hyperparameter.

The idea of the gesture recognition is, similarly to the view estimation, based on a purely geometrical approach and requires that both of the user's arms are visible. According to the elbow angle $\alpha$, we classify for each arm the angle state $s_\alpha\in$ \{\text{Straight}, \text{Perpendicular}, \text{None}\}. In practice, it is not feasible to require the elbow angle to be exactly at $\alpha=90$\textdegree \ for a perpendicular state or at $\alpha=180$\textdegree \ for a straight state. Therefore, we allow a certain interval of degrees, in which the state is recognized as perpendicular or as straight. 

\begin{subnumcases}{ s_{\alpha} =}
         \text{Perpendicular}, & if $\alpha \in (60\text{\textdegree}, 120\text{\textdegree})$ \\
         \text{Straight}, &  if $\alpha \in (140\text{\textdegree}, 220\text{\textdegree})$\\
         \text{None}, & else
    \end{subnumcases}

\begin{figure}[b]
    \centering
    \includegraphics[width=1.0\linewidth]{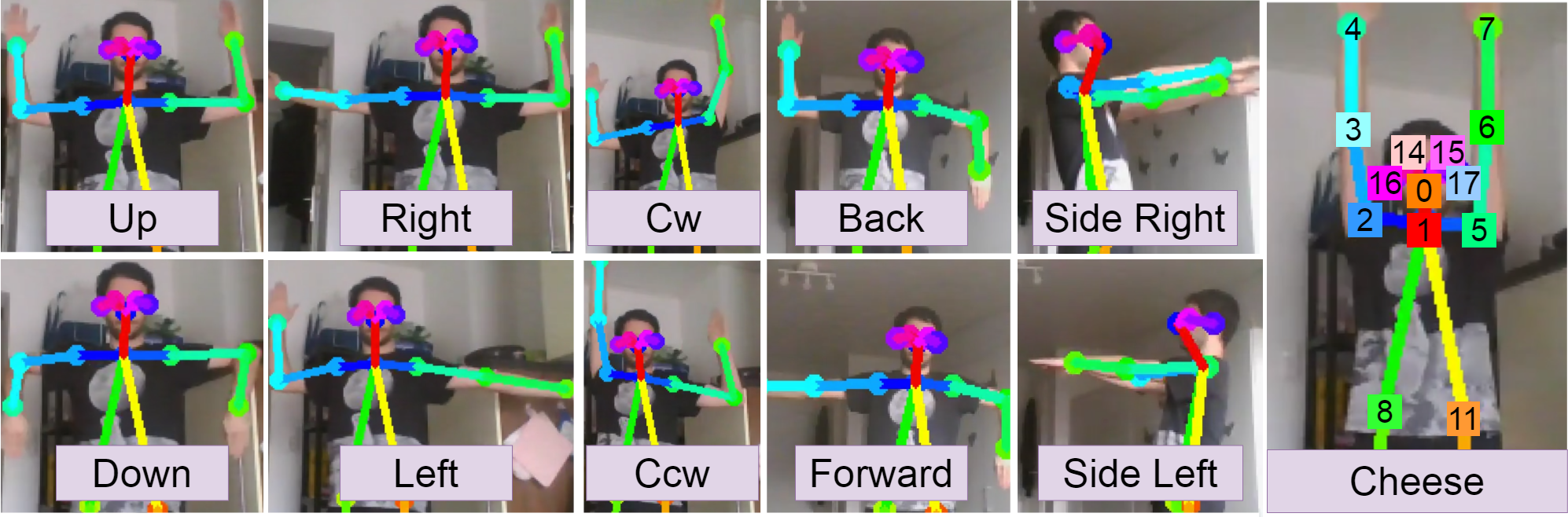}
    \caption{Examples for all the implemented gestures in our framework. The joint labels can be seen in the "Cheese" gesture.}
    \label{fig:all_gestures}
\end{figure}
\begin{figure}[t]
    \centering
    \includegraphics[width=1.0\linewidth]{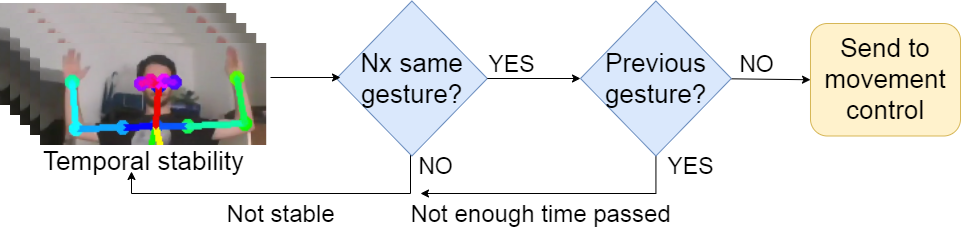}
    \caption{Temporal stability of the recognized gestures}
    \label{fig:temporal_stability}
\end{figure}

Additionally, we also distinguish the position states $s_y\in$ \{\text{Over}, \text{Under}, \text{Middle}\}, which refer to the hand's position with respect to the shoulder. This gives us the tuples $(s_\alpha^l, s_y^l)$ and $(s_\alpha^r, s_y^r)$ for the left and right arm. If $s_\alpha^l$ and $s_\alpha^r$ are both not \textit{None}, the tuples are used to classify a gesture. For example, in Fig. \ref{fig:temporal_stability} we see for both $s_\alpha^l$ and $s_\alpha^r$ are Perpendicular and both $s_y^l$ and $s_y^r$ are Over, which constitute the "Up" gesture. As seen in Fig. \ref{fig:all_gestures} not all possible combinations of angle states and position states are implemented, but the gesture set could easily be extended. 

The side gestures are special gestures, which require the arm to be perpendicular to the body and the view class to be $v_{\text{Side}}$. We included these two gestures so that the user can issue the "Circle Around" (see Fig. \ref{fig:drone_in_use}) command, where the drone flies in front of the user, following a quarter-circle trajectory. We distinguish between "Side Left" and "Side Right" so the drone correctly flies in front of the user, instead of behind him.

A gesture is only forwarded to the movement control when it has been recognized for at least $N$ consecutive frames. If the previously sent gesture is the same gesture, a certain time period $t$ must first pass to avoid sending the same gesture too often as illustrated in Fig. \ref{fig:temporal_stability}. A summary of all implemented gestures can be seen in Fig. \ref{fig:all_gestures}.

\subsection{Distance Module}
The distance module is responsible for estimating the distance to the user. We use the dimensions of the face bounding box $W,H\in\mathbb{R}$ as well as their product $W\times H\in\mathbb{R}$ as input features as their values are distance dependent. We also include the neck-to-waist distance $d(1, 11)\in\mathbb{R}$ for the same reason. Lastly, a one-hot-encoding $\hat{v_i}$ of the estimated view class $v_i\in\mathbb{R}^3$ is used, where $v_\text{Back}$ and $v_\text{Ambiguous}$ are combined into one common class. Then $W,H,W\times H,d(1,11)\text{ and }\hat{v_i}$ are concatenated into a 7-dimensional input vector for the distance estimation. The area $W\times H\in\mathbb{R}$ is a vital feature for the distance estimation, thus we represent it explicitly as input instead of learning it from $H$ and $W$ implicitly.

\begin{figure}[t]
\begin{center}
   \includegraphics[width=1.0\linewidth, keepaspectratio]{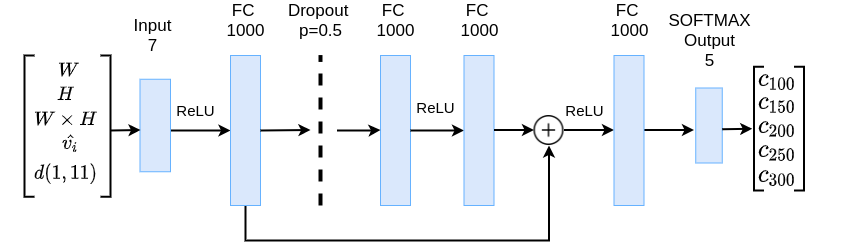}
\end{center}
   \caption{Distance Estimation FCNN Architecture}
\label{fig:distance_FCNN}
\end{figure}

The distance estimation is formulated as a classification problem with the target distance classes $c_i\in\{c_{100},c_{150},c_{200},c_{250},c_{300}\}$, where the indices represent a distance in centimeters. We apply a fully-connected neural network (FCNN) to solve this classification problem as seen in Fig. \ref{fig:distance_FCNN}. The FCNN contains 4 hidden layers with 1000 hidden units each. A dropout layer is added to avoid overfitting and a residual connection, which proved to decrease the training time. The output is a softmax layer, which models a posterior distribution of the distance classes. The training and evaluation of this model are discussed in Section \ref{sec:Eval}.

%
%
To achieve a continuous estimation from the FCNN, for each distance class $c_i$ and its corresponding softmax output $s_i$, a weight $w_i$ is computed by
    \begin{subnumcases}{w_i =}
         s_i, & if $|s_{\text{max}} - s_i| < 0.1$ \\
         0, & else
    \end{subnumcases}
where $s_{\text{max}}$  is the largest softmax output. The resulting weight vector \textbf{w} is then normalized to \textbf{w$'$} so that the sum of the weights is equal to 1. Finally, the weighted sum $\hat{d}=\sum_i c_i w'_i$ is computed, which is a convex combination of the class labels. After the post-processing, the continuous estimated distance $\hat{d}$ is forwarded to the movement control so that the drone can keep a safe distance from the user. 

%
%

\subsection{Movement Control Module}
To ensure the experimental process and provide a variety of control methods, we divide the movement control module into three modes: keyboard, face tracking, and gesture control mode. The distance control is used for the face tracking mode. The keyboard mode is used to switch between modes and to ensure the safe flight of the drone. It can also execute an emergency shut down to avoid collisions. The inputs for the face tracking mode are the distance to user $\hat{d}$ and the center coordinates of the face bounding box $(x,y)$. The face tracking mode sends movement commands with speed and direction information to the drone so that the user’s face is kept in the center of the frame and a certain distance is maintained away from the user. When the gesture control mode receives a gestures command, (\textit{e.g.}, "Up"), from the gesture module, it enables the drone to execute the demanded action.

To control the drone, PID controllers \cite{PID} have to be created for each of these movements. The PID controller is a control loop feedback mechanism that computes the deviation between a given value (measured process value) and the desired value (setpoint) and corrects it based on the proportional, derivative, and integral terms. Equation \eqref{PID_formula} illustrates this.
\begin{equation}
    u(t)=K_p e(t) + K_i \int^t_0 e(t) dt + K_d \frac{de(t)}{dt}
    \label{PID_formula}
\end{equation}
where $e(t)$ is the difference between $r(t)$ and $y(t)$. $r(t)$ represents the setpoint, $y(t)$ the measured process value, and $u(t)$ the control signal. Each PID controller has to be tuned independently to find the correct parameters that will lead to a smooth movement. With Trial and Error method\cite{PIDtuning}, we start with setting both $K_i$ and $K_d$ to zero and increase $K_p$ until the system reaches an oscillating behaviour. Afterwards, we adjust $K_i$ to stop the oscillation and finally adjust $K_d$ for a faster response.

Additionally, in order to complete the gesture control, we need to ensure that the drone can automatically fly to the front of the user. We designed the "Circle Around" control as seen in Fig \ref{fig:drone_in_use}(b), which consists of two actions: fly from the side to the front of the user and rotate 90\textdegree \ to face the user.
%


\section{Evaluation}
\label{sec:Eval}
\subsection{Implementation Details}
\label{subsec:implementation}

\subsubsection{Preliminaries} We used the DJI Tello Drone\cite{tello} to implement our framework. The Tello drone is a quadrotor, which has a 5MP 720p RGB frontal camera and a distance sensor on its bottom side to stabilize its flight. The lack of a frontal depth sensor and a Bluetooth interface encouraged us to implement a monocular image-based distance estimation to the user. To control the movement of the drone and receive information about its state, we utilized the Tello SDK\cite{tello}, which allows connecting to the drone via a WiFi UDP connection. Through this connection, text commands for movement control can be sent as well as queries about the current state of the drone, \textit{e.g.} the current altitude. We use a Python implementation of DLIB\cite{face-recognition} as CNN-based face recognition. The distance module FCNN is implemented with PyTorch and we use a Tensorflow implementation of OpenPose. We utilize an Nvidia GeForce RTX 2070 8 GB and 16 GB of RAM to run our framework with 8-9 FPS. We set $\gamma=0.5$, $N=3$ and $t=625$ms for our hyperparameters. 

\subsubsection{Datasets and training} To perform the supervised training of the FCNN distance estimator, we create a training dataset consisting of 960 images of each distance class $c_i$ (4800 samples in total), where a person performs various gestures to simulate a real-use scenario. We train the network for 15 epochs with softmax-cross-entropy. Additionally, to evaluate our modules we created a custom test dataset, which consists of a person performing all the gestures while rotating to simulate different views and altering the lighting. In all datasets, the gestures are performed with slight modifications concerning the angle and position of the arms. 600 frames are recorded for each distance class, which leads to 3000 test samples in total.

\subsection{Experiments}

\subsubsection{Distance evaluation} The distance estimation is evaluated via the Mean Absolute Error (MAE) and Mean Standard Deviation (MSD) metric in centimeters. The MAE for all the distance classes is estimated at $17.75\text{cm}$ with an MSD of $19.6\text{cm}$. The farthest distance classes contribute to a larger MAE, \textit{e.g.}, for $c_{300}$, the MAE is $30.94\text{cm}$ as seen in Table \ref{tab:distance}.

\begin{table}[b]
\caption{Results of the Distance Estimation Evaluation}
\begin{center}
\begin{tabular}{|c|c|c|c|c|c|c|}
        \hline
            \diagbox[width=0.24\linewidth]{Metric}{Class}
          & $c_{100}$ 
          & $c_{150}$ 
          & $c_{200}$ 
          & $c_{250}$ 
          & $c_{300}$ 
          & \textbf{Average} \\
         \hline
         MAE (cm) & 2.47 & 15.15 & 13.27 & 12.99 & 30.94 & 17.75 \\
         \hline
         MSD (cm) & 5.22 & 12.73 & 15.98 & 11.82 & 24.01 & 19.6\\
        \hline
    \end{tabular}
   \label{tab:distance}
\end{center}
\end{table}

\begin{figure}[t]
    \centering
    \includegraphics[width=\linewidth, keepaspectratio]{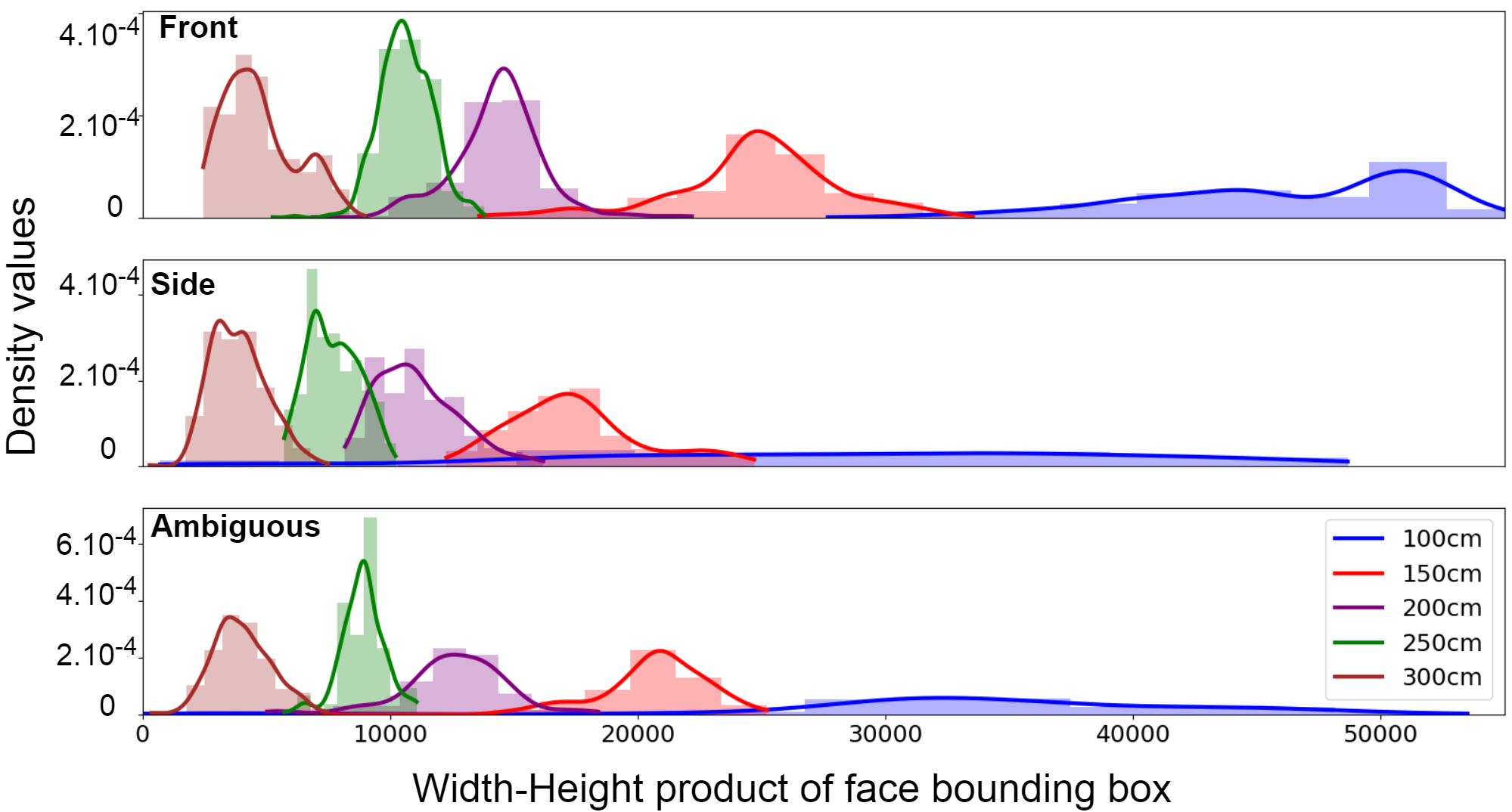} 
    \caption{Distribution of the width-height product in the training dataset.}
    \label{fig:distance_distribution}
\end{figure}

In Fig. \ref{fig:distance_distribution} we can see that the distributions of the width-height product $W\times H$ in our training dataset have a distinct shape with slightly overlapping regions for all distance classes. The distinct shapes of the distributions are a motivation for the FCNN classifier, which is able to separate the distance classes in the latent space. We also see why the one-hot-encoding of the view is needed as an input feature. If the view was not included, the distributions in Fig. \ref{fig:distance_distribution} would have a significant overlap, \textit{e.g.}, $200\text{cm}$ in the side and $250\text{cm}$ in the front view.

%

\subsubsection{Gesture evaluation} As the gesture recognition is entirely based on a geometrical approach, it does not require any training. The only hyperparameter is $N$ which is responsible for the temporal stability of the gestures. It can be set higher if a preciser control is required or lower for a smaller latency. We use the evaluation dataset described in Subsection \ref{subsec:implementation}. In our experiment we consider each frame as stable, \textit{i.e.,} $N=1$. The accuracy of a gesture is equal to the proportion of its correctly recognized frames from the dataset. The proportion of the recognized gestures to all frames is seen in Table \ref{tab:gestures}. The average recognition accuracy of the gestures is 93.5\%.

\begin{table}[b]
\caption{Results from the Gesture Recognition Evaluation}
\renewcommand\arraystretch{1.2}
\begin{center}
\begin{tabular}{|c|c|c|c|}
        \hline
        \multicolumn{3}{|c|}{\textbf{Gesture}} & \multirow{2}*{\textbf{Accuracy}}  \\
        \cline{1-3}
        \textbf{\textit{Type}} & \textbf{\textit{Action}} & \textbf{\textit{Name}} &   \\
        \hline
        \multirow{9}*{Frontal} & \multirow{6}*{\shortstack{Translational \\ movement}} & Up & 98.8\% \\  
        &&Down & 66.6\% \\ 
        &&Left & 98.6\% \\ 
        &&Right & 97.1\% \\ 
        &&Forward &  97.4\% \\  
        &&Backward & 94.1\% \\  
        \cline{2-4}
        & \multirow{2}*{\shortstack{Rotational \\ movement}} &Clockwise (Cw) & 97.4\% \\  
        &&Counter-clockwise (Ccw) &  97.2\% \\ 
        \cline{2-4}
        & Take a photo & Cheese & 96.6\% \\  
        \cline{1-4}
        \multirow{2}*{Side} & \multirow{2}*{\shortstack{Circle \\ around}} & Side-left & 89.2\% \\  
        &&Side-right & 95.6\% \\  
        \hline
        \multicolumn{3}{|c|}{\textbf{Average accuracy}} & \textbf{93.5\% }\\ 
        \hline
    \end{tabular}
   \label{tab:gestures}
\end{center}
\end{table}
The gesture accuracy depends entirely on the joints estimations from OpenPose and the correctness of the gesture execution from the user. The gestures are designed in a natural way as each gesture represents the demanded movement, \textit{i.e.,} pointing left for the drone to fly left. As many of the gestures have an opposite gesture, they are mapped to mirrored arm positions as in Fig. \ref{fig:all_gestures}. (\textit{e.g.,} "Up" and "Down" gestures). The only gesture, which is not easily recognized is the "Down" gesture as it is physically difficult to perform.

\section{Conclusions}
The results from the evaluation show that despite the hardware limitation of the DJI Tello drone, it is possible to achieve a robust distance estimation as well as a reliable gesture recognition. This is made possible by leveraging a skeleton-based pose estimation, which we showed provides sufficient features for both tasks. The high average gesture recognition accuracy (93.5\%) enables a smooth and natural HDI and the low MAE/MSD for the distance estimation ensures the user's safety. The framework can control the drone in real-time, enabling a natural and simple interaction. Due to its modularity and extensibility, our framework could be integrated and further developed in various other applications.

\bibliographystyle{ieeetr}
{\small\bibliography{refs}}

\end{document}